# A two-stage learning method for protein-protein interaction prediction

[1]Amir Ahooye Atashin, [2]Parsa Bagherzadeh, [3]Kamaledin Ghiasi-Shirazi

Department of Computer Engineering
Ferdowsi University of Mashhad
Mashhad, Iran
[1]amir.atashin@stu.um.ac.ir
[2]parsa.bagherzadeh@stu.um.ac.ir
[3]k.ghiasi@um.ac.ir

*Abstract*— In this paper, a new method for PPI (protein-protein interaction) prediction is proposed. In PPI prediction, a reliable and sufficient number of training samples is not available, but a large number of unlabeled samples is in hand. In the proposed method, the denoising autoencoders are employed for learning robust features. The obtained robust features are used in order to train a classifier with a better performance. The experimental results demonstrate the capabilities of the proposed method.

*Protein-protein interaction; Denoising autoencoder; Robust features; Unlabelled data;*

I. INTRODUCTION

Protein-protein interaction (PPI) is one of the most important processes in biological systems. PPI refers to physical contacts established between two or more proteins. These interactions often perform biological functions. Many of important molecular processes in the cell are performed by large molecular complexes which are consist of a large number of interacting proteins. Because of their importance for understanding the biological functions, prediction of PPI's is the subject of research and has attracted great attention.

The prediction of PPI is performed either experimentally or computationally. Experimental methods are consisting of biochemical and genetic experiments for PPI prediction. Most of these methods are expensive and very time-consuming. Yeast two-hybrid screening and Mass spectrometry are examples of these methods [1]. On the other hand, in the past years, investigation of PPI using computational methods has become the subject of research interest[2]. Computational methods often refer to the deployment of machine learning techniques to obtain a predictive model for PPI prediction. These models are built based on known pairs of protein which are labeled as interacting or non-interacting. In other words, the goal of computational methods is the integration of various source of evidence in a statistical framework.

In computational methods, the goal is the learning of distribution $P(Y|X)$ from the available instances. This problem is classification and a supervised learning task in which the distribution is learnt form the set of $n$ instances $\{(x_1, y_1), ..., (x_n, y_n)\}$ associated with predefined labels. It is assumed that the instance of this set are independent samples of the actual distribution $Q(Y|X)$. Several examples of computational methods are naive logistic regression [3], random forest based method [4], Bayes classifier [5], decision tree [6], kernel based methods from [7], [8], and the strategies of summing likelihood ratio scores [9-11].

Despite their advantages, the main drawback of computational methods is the availability of reliably labeled samples. A number of these samples are not sufficient for the construction of a comprehensive model. On the other hand, a large number of unlabeled or partially labeled instances are available. It is indicated empirically that a feature vector which is more robust to noise can improve the performance of the classifier. Obtaining such set of features can be an initial step for the main classification problem. These kinds of features can be obtained from an expert, but the main drawback of this approach is its cost and time consumption.

A less expensive and more time-saving approach can be obtained using Denoising Auto-Encoder (DAE) [12]. DAE is an unsupervised learning model which tries to achieve a useful representation of data. Using this method, a set of robust feature can be extracted automatically from data. The Denoising Auto-Encoder (DAE) is an extension of a classical autoencoder and it was introduced as a building block for deep networks [13]. Since the DAE is an unsupervised learning model, it can benefit from a large number of unlabeled or partially labeled pairs of proteins and can be employed for solving the PPI prediction problem [14].

In this paper, a new method for solving PPI prediction problem, based on denoising autoencoder is proposed. A Denoising Auto-Encoder is employed for learning of robust features. The learned features are used for training a multi-layer feed forward neural network with a better performance [15].

The remainder of this paper is organized as follows: section II presents some preliminaries concerning autoencoder, denoising autoencoder and stacked autoencoder [16]. In section III the proposed method for PPI prediction is presented. Section IV reports the experimental results and finally section V gives the concluding remarks.

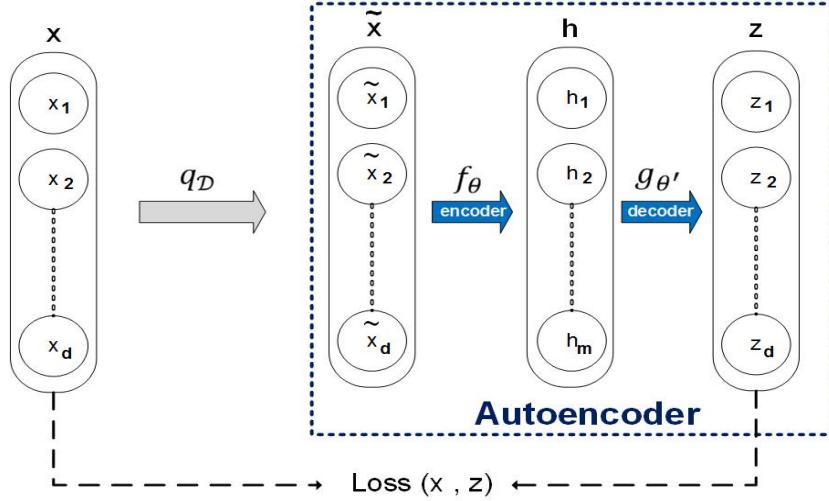

Figure 1. Illustration of denoising autoencoder network. A particular sample $x$ is corrupted to noisy sample $\tilde{x}$ via $q_D$. Then autoencoder maps $\tilde{x}$ to $h$ with encoder $f_\theta$ and try to reconstruct orginal sample $x$ from $h$ with decoder $g_{\theta'}$. The objective of the DAE is minimizing reconstruction error that computed by loss function ( $Loss(x,z)$ ).

## II. PRELIMINARIES

*In this section, the denoising autoencoder is reviewed. We begin with a short discussion on Autoencoders.*

### A. Auto-Encoder (AE)

An autoencoder can be described as two-part, encoder, and decoder. The Encoder is a function $f$ that takes an input $x \in [0,1]^d$ and maps it to hidden representation $y \in [0,1]^{d'}$ through a deterministic mapping.

$$y = f(x) = S(Wx + b_y) \quad (1)$$

Where S is a nonlinear function such as a sigmoid, and W is weight matrix with a size of d x d'.

The Decoder is a function $g$ that takes a representation y back to reconstruction z:

$$z = g(y) = S(W'x + b_z) \quad (2)$$

In this paper, we explore the tied weight case, that $W' = W^T$.

The autoencoder training can be done by finding parameters $\theta = \{W, b_y, b_z\}$ that minimized the reconstruction error on a training set of data $X$.

$$J(\theta) = \sum_{x \in X} L(x, g(f(x))) \quad (3)$$

The reconstruction error can be measured in many ways. Typically choice is *squared error* $L(x,y) = \|x-y\|^2$ or the *cross-entropy loss* when the input is interpreted as either bit vectors or vectors of bit probabilities:

$$L(x,y) = \sum_{i=1}^{d} [x_i \log z_i + (1-x_i)\log(1-z_i)] \quad (4)$$

### B. Denoising Auto-Encoder (DAE)

The denoising autoencoder is a stochastic version of the autoencoder, where one simply corrupts the input $x$ before sending it to the autoencoder, which is trained to reconstruct the clean data (to denoise). This yields the following objective:

$$J(\theta) = \sum_{x \in X} \mathbb{E}_{\tilde{x} \sim q(\tilde{x}|x)} \left[ L(x, g(f(\tilde{x}))) \right] \quad (5)$$

Where the expectation is over corrupted versions $\tilde{x}$ of examples x obtained from a corruption process $q(\tilde{x}|x)$. This objective is optimized by stochastic gradient descent. In this paper corruption is considered as additive isotropic Gaussian noise: $\tilde{x} = x + \epsilon, \epsilon \in \mathcal{N}(0, \sigma^2 I)$ and a binary masking noise where a fraction $\nu$ of input components have their value set to 0.

### C. Stacked Denoising Auto-Encoder (S-DAE)

Denoising autoencoders can be stacked to form a deep network by feeding the hidden representation of the denoising autoencoder found on the layer below as input to the current layer. The unsupervised pre-training of such an architecture is done one layer at a time. Each layer is trained as a denoising autoencoder by minimizing the error in reconstructing its input (which is the output code of the previous layer). Once the $i$th layer is trained, we can train the $i + 1$th layer because we can now compute the hidden representation from the previous layer.

Once all layers are pre-trained, the next step is to train network on the supervised manner, this phase called fine-tuning. Here we consider fine-tuning where we want to minimize prediction error on a classification task.

## III. THE PROPOSED METHOD

In PPI prediction a small set of labeled data $D_l = \{(x_1, y_1), ..., (x_n, y_n)\}$ where $x_i \in \mathcal{R}^D$ is i-th data and with corresponding class label $y_i \in \{0,1\}$, and we have a large set of unlabeled data set $D_u = \{(x_1, ..., x_m)\}$ are in hand.

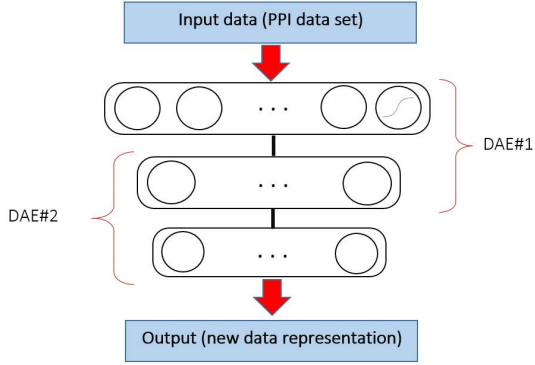

Figure 2. An example of a stacked DAE with two layers

Thus, we are facing with a semi-supervised classification problem. Semi-supervised learning is a class of supervised learning methods that not only uses the labelled instances for training but also make use of unlabeled data.

Now we develop an algorithm for semi-supervised learning that employs the unlabeled data to improve the classification performance. DAE is used to construct a model that is able to provide a more robust set of features from data. The proposed model is consisting of stacking several layers of DAEs. Each of these DAEs is employed to build a robust feature vector from its input by learning original inputs from their corresponding corrupted versions.

Firstly, each DAE is trained based on an unsupervised layer-wised greedy algorithm in which each layer is trained separately, one by one from bottom to up. Since DAE trained via unsupervised learning, it tries to find dependency between input variables. Since adding noise correspond to regularization [15] the obtained representation at final layer leads to a better generalization capability.

At the final stage, we build a classifier based on the output signal of the s-DAE model. Since the input of the classifier is a more robust version of the original input, it is expected to have a more generalization for test data.

## IV. EXPERIMENTAL RESULT

In this section, the performance of proposed method is evaluated on PPI prediction problem.

### A. Dataset

The proposed method is applied to HIV-1 and human protein-protein interaction prediction problem [17]. This data set is consisting of labeled and unlabeled pairs. Both parts are used in the pre-train phase. The pairs of proteins are described in 18-dimensional space. Interacting pairs of proteins are considered as positive training samples. On the other hand, non-interacting pairs are considered as a negative class. The interaction of these labeled samples is indicated by HIV experts computationally and the second part of the dataset consists of unlabeled pairs of proteins. For these unlabeled pairs, there is not enough evidence for indication of interaction so they are left unlabeled. The characteristics of the dataset are summarized in table I.

### B. Model architecture

*1) Unsupervised pre-training:* Two stacked denoising autoencoder are used. These denoising autoencoders are chosen to be contrastive. The first DAE tries to map the 18-dimensional input vector to 14 dimensions. The second DAE maps the output of first DAE to 8 dimensions. This 8-dimensional feature vector is the robust and generalized version of 18-dimensional input feature vector. The stacked DAE model is trained layer wised. The parameters of Sacked DAE are summarized in Table II. A visual representation of the filters learned by denoising autoencoders is shown in Figure 3. As it can be seen a meaningful representation of features is obtained.

TABLE I. CHARACERISCICS OF HIV-1 AND HUMAN PPI PREDICION

| *Features* | *Positive PPIs* | *Partial positive* | *Remaining pairs* |
|---|---|---|---|
| 18 | 158 | 2188 | 352338 |

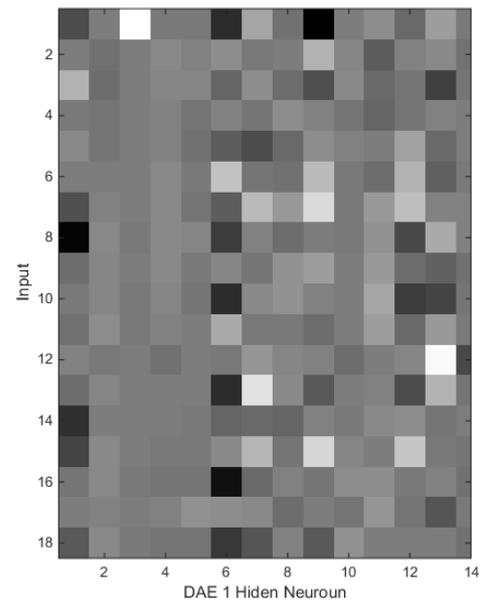

(a)

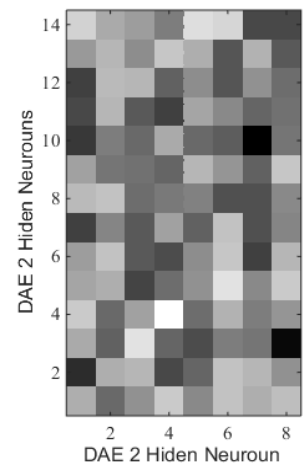

(b)

Figure 3. Visualization of the filters learned by denoising auto encoders. each column shows the weights connecting a hidden neuron to the inputs from previous layer. The weights magnitued scaled to 0 to 255 for visulizations.

TABLE II. PARAMEERS OF S-DAE MODEL

| DAE level | #Hidden neurons | Input noise fraction | Learning rate | momentum |
|---|---|---|---|---|
| 1 | 14 | 0.4 | 1 | 0.1 |
| 2 | 8 | 0.1 | 0.5 | 0.1 |

*2) Supervised training:* A three layer feed-forward network is constructed based on the obtained weights of unsupervised phase. The first two layers are initialized by the sDAE weights and the weights of the third layer are assigned randomly. The output of the feed-forward network is a single sigmoid neuron, indicating the respective class of the input based on the lowest difference of the output to either 1 or 0. The feed forward network is trained using back propagation algorithm by labeled samples.

TABLE III. PARAMEERS OF FEED-FORWARD NEURAL NET TRAINING

| Architecture | Activation function | #epochs | L2 weight penalty | Learning rate | momentum |
|---|---|---|---|---|---|
| 18-14-8-1 | sigmoid | 2000 | 0.0007 | 1 | 0.5 |

*C. Numerical Results:* In order to demonstrate the capabilities of the proposed method for PPI prediction, it is compared against other classical classification methods such as SVM, kNN, and MLP. The experiments were performed using the k-fold approach with $k = 5$. The accuracies of PPI prediction, obtained by each of these classifers are shown in Table IV.

TABLE IV. OBTAINED RESULT OF PPI PREDICTION USING CLASSICAL CLASSIFICATION METHODS AND THE PROPOSED METHOD

| Predictor | Accuracy | Precision | Recall |
|---|---|---|---|
| kNN | 40.3% | 0.32 | 0.43 |
| SVM | 52.6% | 0.48 | 0.51 |
| MLP | 64.4% | 0.57 | 0.65 |
| Porposed method | **73.8%** | 0.70 | 0.77 |

## V. CONCLUSIONS

In this paper, we proposed a new method for protein-protein interaction prediction problem. Our two-stage learning method used a large number of unlabeled data to extract robust features from input data and in the second phase, we applied them to achieve better performance on the classification stage of our method. For future work, we plan to see our method as learning framework which other generative models like Restricted Boltzmann Machine or discriminative models like support vector machine, can be replaced and tested in the first and second stage of our method.

ACKNOWLEDGMENT

The authors would like to thank Dr. Tasan et al. for providing the Human Protein to HIV-1 Virus Interaction Prediction data set and Dr. Qi et al. for providing the "Gold standard" positive samples.